\documentclass[9pt,twocolumn,twoside]{pnas-new}
\usepackage{cleveref}

\templatetype{pnasresearcharticle} % Choose template 
\usepackage{graphicx}
\usepackage{subcaption}

\title{Culturally-Attuned Moral Machines: Implicit Learning of Human Value Systems by AI through Inverse Reinforcement Learning}

\author[a]{Nigini Oliveira}
\author[a]{Jasmine Li}
\author[a]{Koosha Khalvati}
\author[b]{Rodolfo Cortes Barragan}
\author[a]{Katharina Reinecke}
\author[b]{Andrew N. Meltzoff}
\author[a, 1]{Rajesh P. N. Rao} 

\affil[a]{Paul G. Allen School of Computer Science and Engineering, University of Washington, Seattle, USA}
\affil[b]{Institute for Learning and Brain Sciences, University of Washington, Seattle, USA}

\leadauthor{Oliveira} 

\significancestatement{
One of the most important questions facing the integration of artificial intelligence (AI) into human society today is: how can AI systems  be imbued with the diverse moral and ethical norms of human cultures across the world?  Rather than incorporating a fixed universal moral code, we argue that AI systems should emulate how children learn values and norms through observation of others within their specific culture. By applying inverse reinforcement learning to human behavioral data, we illustrate how AI systems can learn culturally-attuned value systems implicitly by observing the behavior of humans from different cultural groups.}

\authordeclaration{The authors declare no competing interests.}
\correspondingauthor{\textsuperscript{1}To whom correspondence should be addressed. E-mail: rao@cs.washington.edu}


\begin{abstract}
Constructing a universal moral code for artificial intelligence (AI) is difficult or even impossible, given that different human cultures have different definitions of morality and different societal norms. We therefore argue that the value system of an AI should be culturally attuned: just as a child raised in a particular culture learns the specific values and norms of that culture, we propose that an AI agent operating in a particular human community should acquire that community's moral, ethical, and cultural codes. How AI systems might acquire such codes from human observation and interaction has remained an open question. Here, we propose using inverse reinforcement learning (IRL) as a method for AI agents to acquire a culturally-attuned value system {\em implicitly}. We test our approach using an experimental paradigm in which AI agents use IRL to learn different reward functions, which govern the agents' moral values, by observing the behavior of different cultural groups in an online virtual world requiring real-time decision making. We show that an AI agent learning from the average behavior of a particular cultural group can acquire altruistic characteristics reflective of that group's behavior, and this learned value system can generalize to new scenarios requiring altruistic judgments. Our results provide, to our knowledge, the first demonstration that AI agents could potentially be endowed with the ability to continually learn their values and norms from  observing and interacting with humans, thereby becoming attuned to the culture they are operating in.
\end{abstract}

\begin{document}

%\flushbottom
\maketitle
\thispagestyle{firststyle}
\ifthenelse{\boolean{shortarticle}}{\ifthenelse{\boolean{singlecolumn}}{\abscontentformatted}{\abscontent}}{}

\section*{Introduction}
A formidable challenge in artificial intelligence (AI) today is developing AI models that can incorporate the diversity of human culture in their reasoning, bringing with it the richness and depth of ethical and moral considerations. In this article, we hypothesize that rather than attempting to handcraft a universal moral code for AI, AI agents could implicitly learn cultural rules, morals, norms, and values by being embedded in a human culture. Such an approach would make these AI agents culturally-attuned moral machines, endowing them with the ability to deal with issues of morality, inter-individual rights and obligations, and decisions about one's responsibilities toward others in a manner that is tailored to the culture the AI agent was ``raised in.'' This acknowledges the fact that what is morally acceptable in one culture may not be acceptable in another \cite{oshotse_2021_cultural}, necessitating AI systems that are capable of adapting to the moral codes of the culture in which they are operating. This is especially important as even humans often have a hard time appreciating the values of other cultures \cite{Kunst_Acculturation_2021, oshotse_2021_cultural}. 

Our approach is inspired by evidence regarding how neural mechanisms in the brain support social and moral behavior: the same reward systems thought to be involved in reinforcement-based learning are also harnessed to learn the dominant norms and values within a society~\cite{dolen2013social, hung2017gating,churchland_conscience_2019}. The resulting system of learned rules governs a human's cultural expectations and is arguably the basis of social values and moral thinking. Thus, to learn what the accepted norm is in any given context, the brain must first develop expectation rules about events and learn internal value systems, including the benefits of certain behaviors as well as the costs of deviating from the norm. For example, by observing and participating in sharing behaviors within a culture, a child or an adolescent may learn to associate, in a certain context, a `positive' value with behaving altruistically \cite{flores2018adolescents}.

Although there has been considerable AI research on using inverse reinforcement learning and imitation learning to mimic human actions \cite{Ng-IRL00,Verma-Rao-05,Rao-Meltzoff-07,Abbeel-Ng-04,Hussein-imitation-learning-2017,Adams-survey-IRL-2022}, we are not aware of any AI research on uncovering the internal `value systems' of human behavior, and particularly on learning moral codes from observing human actions within a culture. Here, we focus on this issue by asking: \textit{Can an AI agent learn culturally-attuned quantitative value systems for moral and ethical scenarios by observing the behavior of humans from different cultural groups?} We expect the knowledge gained from learned value systems to contribute to the AI's explainability, i.e., transparency regarding why an AI model produced a certain output, and to allow the AI to extrapolate its learned moral and ethical codes to novel situations. Such AI systems could also be used to understand human behaviour and human value systems in complex naturalistic settings. 

In this article, we leverage the technique of Inverse Reinforcement Learning (IRL)~\cite{ng_irl_2000,arora_irl_2021} to allow an AI agent to infer and learn the reward functions of humans belonging to a particular community or cultural group. The learned reward function assigns specific reward values (positive or negative) to situations in which the AI agent may find itself. By choosing actions that optimize total expected reward based on its learned reward function, the AI agent can behave in a manner that is attuned to the human culture in which it is embedded. A major advantage of this approach is that if the reward function is learned using a function approximator (e.g., neural networks as in our implementation), the agent can assign values to moral scenarios that are similar to but not exactly the same as previously encountered scenarios, allowing the agent to generalize its moral decision making ability to novel situations, as children do based on cultural learning and experience \cite{barragan_possessive_2021,barragan_2024}. 

To test this approach, we collected data from human participants playing an online game involving altruistic choices and recruited participants with cultural backgrounds that would likely differ in altruistic behavior, specifically, participants who identified as White Americans and Latino Americans. Our hypothesis regarding differences in altruistic behavior is motivated by the extensive cross-cultural literature showing that Latino Americans tend to be more collectivistic~\cite{krys_latinamerica_2022,triandis_individualism_1995} and have a more interdependent self-concept than White Americans, who are generally thought to be more individualistic and prone to seeing  themselves as independent of others~\cite{markus_interdependence_1991, salvador_interdependence_2023,kitayama_2024}. Whereas individualistic cultures tend to emphasize personal achievement and autonomy, collectivist cultures prioritize the achievements and welfare of others over personal gains. We  hypothesize that Latino Americans will exhibit more altruistic behavior in our study than White Americans. To control for other societal characteristics, we decided only to recruit individuals who live in the USA, and who self-declared as `Latino' or `White'. Finally, our online game scenario has strong ties to ``morality'' as it emulates a situation involving ``unfair'' treatment of an individual \cite{levine2020logic}. In the scenario, the ``privileged'' player has to endure some costs to help the other player who has been placed in a disadvantaged and potentially less rewarding circumstance.    

Rather than using data from human adults answering questions about moral scenarios, as in the case of the Moral Machines experiment~\cite{awad_moral_2018}, or data from descriptive judgments of ethical and moral scenarios, as in the Delphi experiment~\cite{jiang_delphi_2021}, our approach emulates the way a human child absorbs the ethical and moral values of their culture~\cite{barragan_possessive_2021, meltzoff_imitation_2018}: by observing the actions of others and implicitly learning to assign values to particular situations and behaviors. A key advantage of our approach is that the AI would be able to dynamically adapt to changes in moral values, as expected in a constantly evolving cultural environment. 

%%% Summary of results
Our results show that an AI agent learning from a particular cultural group (Latino or White Americans) can acquire altruistic characteristics reflective of that group's behavior through IRL, and this learned value system can generalize to new scenarios that require altruistic decision making. Our results pave the way for AI agents of the future that remain attuned to the culture they operate in by continually learning values and norms from observing and interacting with humans within that culture.

%Researchers have previously focused on AI agents that learn morals and norms by learning language models \cite{jiang_delphi_2021} or have sought to understand `morality' in machines through human surveys \cite{awad_moral_2018} and by creating Artificial Moral Agents~\cite{zoshak_ethicalAMA_2021}. Our work differs from these previous approaches in leveraging human behavioral data to train AI agents using IRL to implicitly learn reward functions from human actions, producing culturally attuned AI that can potentially generalize to new scenarios.

\section*{Experimental Study}
\begin{figure*}[t]
    \centering
    \begin{subfigure}[b]{0.4\textwidth}
        \centering
        \includegraphics[width=\textwidth]{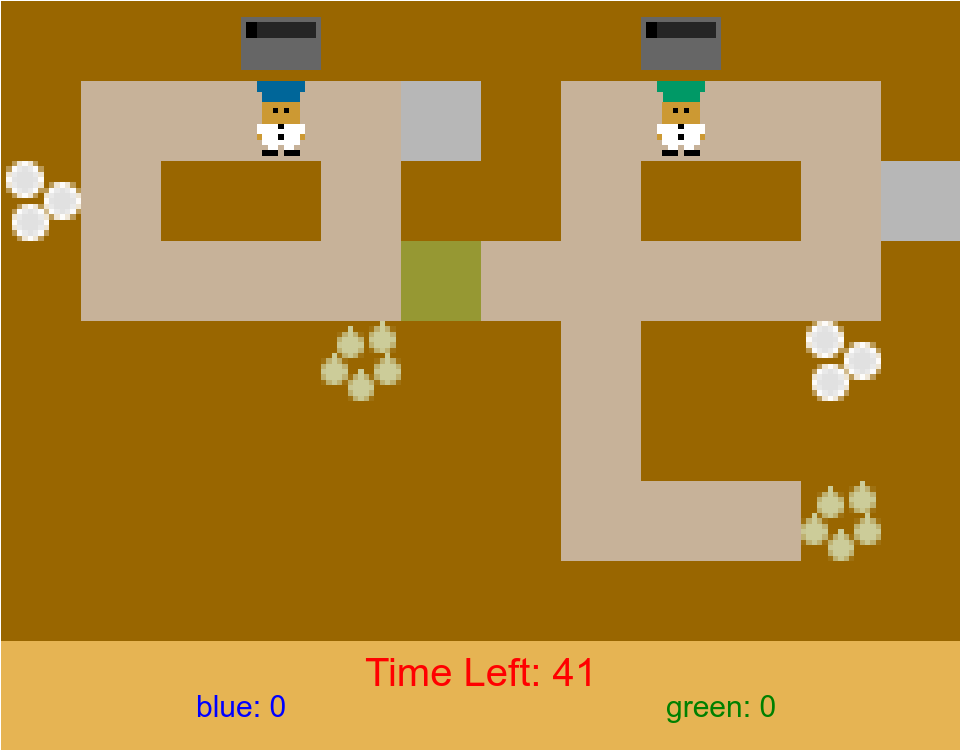}
        \caption[]%
        {}
        \label{fig:game:intro}
    \end{subfigure}
    \hspace{0.2in}
    %\hfill
    \begin{subfigure}[b]{0.4\textwidth}  
        \centering 
        \includegraphics[width=\textwidth]{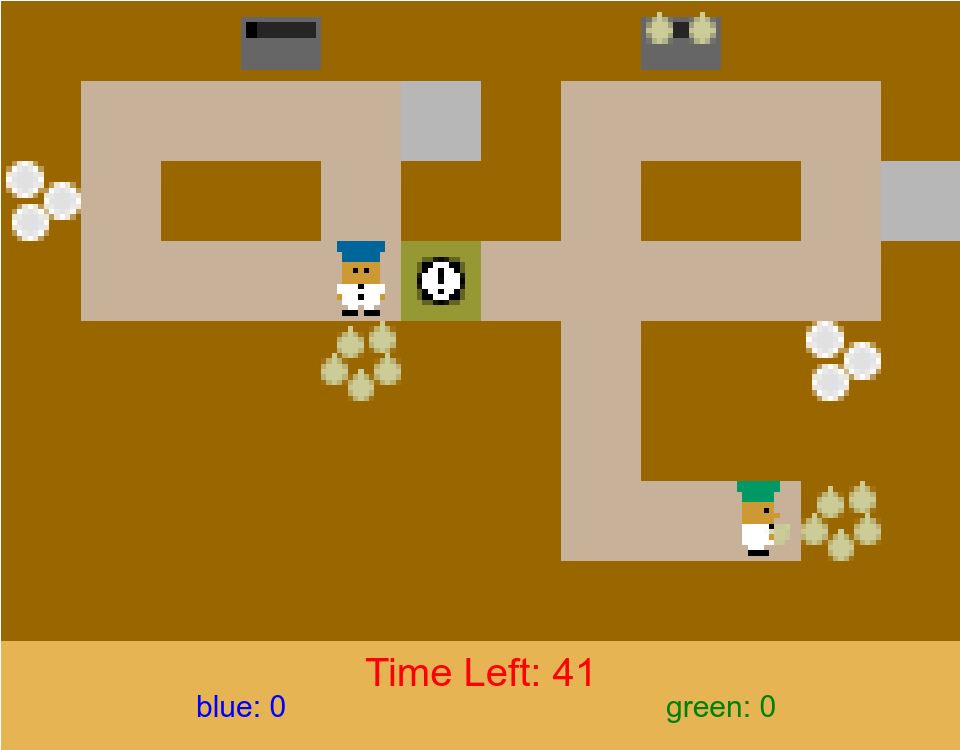}
        \caption[]%
        {}
        \label{fig:game:full}
    \end{subfigure}
%    \vskip\baselineskip
    \caption[]{{\bf Testing Altruistic Behavior using an  Online Experiment}. {\bf (a)} The screenshot shows both players, one in a blue chef’s cap and the other in a green one, in front of their respective stoves (dark gray with black bars inside). The goal is to cook and deliver as many onion soups as possible within a given time limit by putting three onions (yellow ovals with stalks) in a cooking pot on the stove, picking up an empty bowl (white circle), collecting the cooked soup in the bowl from the stove, and delivering the bowl to customers (light grey location). Players have separate scores and can choose to cook soups individually to get points or share onions. Players can share onions by placing them on the ``cooperation bridge'' (square colored green between the two kitchens). One can see how much shorter the path is for the player on the left to get to the (continually replenished) onion store at the bottom of the kitchen compared to the player on the right. Each study participant plays Round 1 (altruistic behavior baseline) on the left side, Round 2 (altruistic/non-altruistic bot demonstration round) on the right side, and Round 3 (behavior change examination round) on the left side again. When the computer-controlled bot is playing on the left in Round 2, it randomly selects one of two possible behaviors: altruistic behavior, where the bot places onions on the cooperation bridge for the human player, or non-altruistic behavior, where the bot does not help and focuses on maximizing its own score. {\bf (b)} In this screenshot, the players are in front of their onion stores. The exclamation icon on the cooperation bridge denotes a call for help left by the player on the right before going all the way to the southeast end to get onions from their own store.}
    \label{fig:game}
\end{figure*}

We trained our AI on human behavior in an online experiment that required subjects to make decisions around altruistic behavior as part of a larger context of actions. Altruism -- or behaving in ways that benefit another individual at a cost to oneself --is desirable in all cultures, but we expect variations based on different norms regarding altruistic behavior across the two cultural groups we recruited in our online experiment. Figure~\ref{fig:game} presents a snapshot from the online experiment. The experiment used a version of the commercially successful ``Overcooked''  game in which players control chefs in a kitchen to prepare meals, given specific orders, within a time limit. We implemented a simplified version of the game building on previous research in AI that used this game to study human-AI coordination~\cite{carroll_utility_2019}.

As shown in Figure~\ref{fig:game}, the game involves two chefs (wearing blue and green hats), each controlled by either a human participant or the computer. The goal is to cook and deliver as many onion soups as possible within a given time limit. This involves putting three onions (yellow ovals with stalks at the top) in a pot on the stove (dark grey), then picking up an empty bowl (white circle), collecting the cooked soup on the stove, and delivering the bowl of soup to customers (light grey location). Before the experiment began, our study taught participants how to play the game through a series of interactive tutorials. The participants also provided basic demographic information. In every round, the participant was paired with a computer-controlled bot on the other side of the kitchen. Importantly, cooking in one side of the kitchen (the right side) required more effort as the chef needed to traverse a longer path to obtain an onion and put it on the stove. This asymmetry created an unfair situation and the chef on the easier side (the left side) could alleviate this unfairness by passing an onion to the other chef through a ``cooperation bridge'' (green square in Figure~\ref{fig:game}). Such an altruistic act, however, hurts the helping chef's performance as it consumes time that could otherwise have been used to deliver more soups. 

In our experiment, participants played three rounds: in Rounds 1 and 3, they controlled the chef on the left side of the kitchen (where the onions are closer to the stove), and in Round 2, they controlled the chef on the right side (where the onions are farther away from the stove).  When the computer was playing on the right side, it was programmed to call for help every time it moved empty-handed past the bridge connecting the two kitchens (denoted by an exclamation mark in Figure~\ref{fig:game:full}). In Round 2, half of the participants were randomly assigned to be paired with a  computer-controlled bot on the left side that demonstrated altruistic behavior by depositing an onion on the bridge for the human-controlled chef. The other half of the participants were randomly paired with a self-serving bot that never shared the onions, resulting in the human player having to go to a more distant location to get onions. This design choice was meant to test the hypothesis that participants who received help in Round 2 might increase their altruistic behavior in Round 3 (helping the chef on the right side by putting onions on the bridge) compared to participants who did not. The overall goal remained the same: to achieve higher scores by cooking as many soups as possible in a 60-second round. The current scores for each player (blue and green chef) and the time left on the clock were shown on the screen (see Figure~\ref{fig:game}). A chef's score, whether human- or computer-controlled, increased by 10 points whenever the chef delivered a soup.

\section*{Results}

In the following, we present two sets of results. The first set of results are from our online experiment. These confirm our hypothesis regarding behavioral differences across cultures pertaining to altruistic decision making. The second set of results shows how these differences can be learned by culturally-attuned AI agents using inverse reinforcement learning.

\subsection*{Altruistic Behavior and Cultural Variations in the Online Experiment}
For the online experiment described above, we recruited 300 adult participants from two groups of US residents: one group self-identified as `White' ($n=190$) while the other self-identified as `Latino' ($n=110$). Data was collected across three major online recruitment platforms: Prolific, Positly, and Amazon MTurk. The final dataset contained participants' summary results per round and all game-playing information, i.e., sequential game-state snapshots, including each board arrangement, actions taken by both agents, time, and scores. Demographic information such as education level and political orientation was also collected (details in the Methods Section).

\begin{figure*}
    \centering
    \begin{subfigure}[b]{0.48\textwidth}  
        \centering 
        \includegraphics[width=\textwidth]{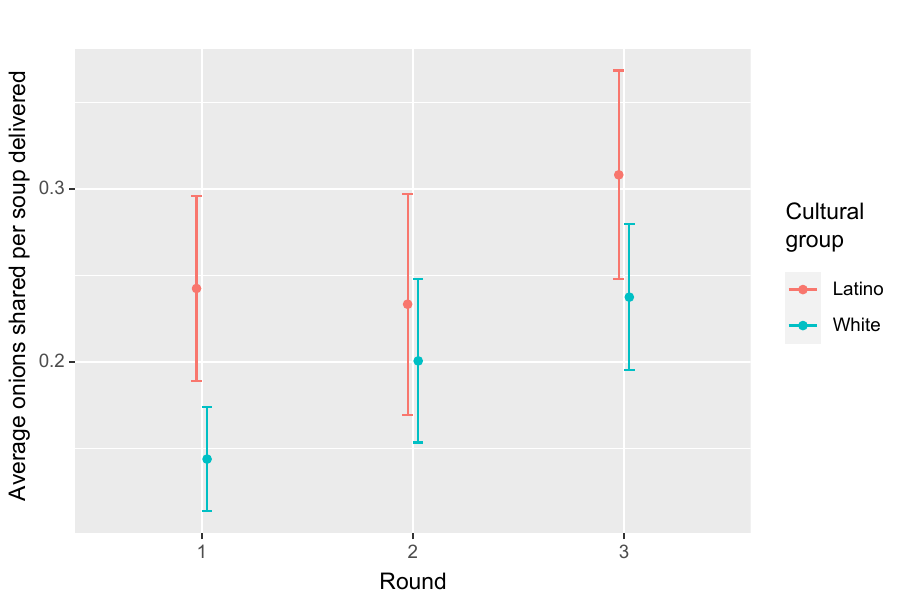}
        \caption[]%
        {}    
        \label{fig:altruism:race}
    \end{subfigure}
        \hfill
    \begin{subfigure}[b]{0.48\textwidth}
        \centering
        \includegraphics[width=\textwidth]{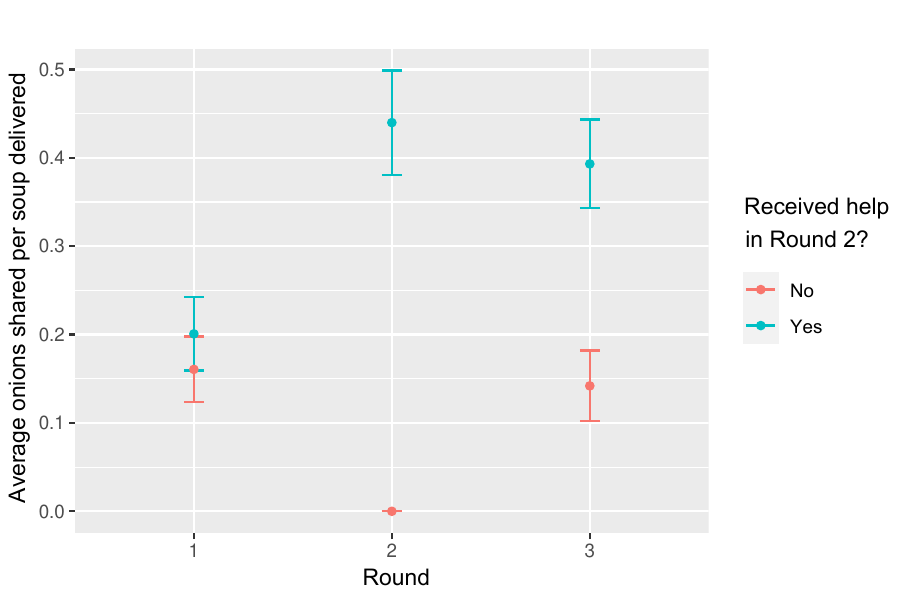}
        \caption[]%
        {}
        \label{fig:altruism:rounds}
    \end{subfigure}
%    \vskip\baselineskip
    %\caption[]{{\bf Cultural Effects on Altruistic Behavior in the Online Experiment}.  
     \caption[]{{\bf Cultural Effects on Altruistic Behavior in the Online Experiment}. In both figures, the Y axis shows the average human altruistic behavior in terms of onions shared per delivered soup. {\bf (a)} shows a tendency for Latino Americans to share more onions, although with statistical significance only in Round 1 (considering a confidence interval of 95\%). {\bf (b)} shows the effects of the computer-controlled bot helping versus not helping the human participant in Round 2: participants who received help in Round 2 demonstrated significantly more altruistic behavior in Round 3. In contrast, those who were not helped maintained approximately the same level of altruistic behavior. } 
    %\label{fig:results}
\end{figure*}

We first tested our hypothesis, motivated by previous research on cooperative tendencies in Latino and White American populations~\cite{triandis_individualism_1995,salvador_interdependence_2023}, that cultural background may influence altruistic behavior in our participants at the outset of the experiments. Figure~\ref{fig:altruism:race} supports this hypothesis: Latino participants were more altruistic at the outset (in Round 1 of the online game) compared to White participants (ANOVA, $F=11.8$ and $\alpha=0.0007$), with Latino participants averaging $0.24$ (SD = $0.28$) shared onions per delivered soup and White participants averaging $0.14$ (SD = $0.21$)) (Cohen’s $d = 0.40$ when comparing Latino and White participants). Our data thus provides an example where moral decision-making, in this case, a specific altruistic behavior, differs across two cultural groups.\footnote{We note that the linear model including gender, age, race, and political orientation only explains 8\% of the variance in altruistic behavior in our dataset.} 

We next examined the hypothesis that participants who receive help in Round 2 may increase their helping behavior in Round 3 compared to those who did not receive help in Round 2. Figure~\ref{fig:altruism:rounds} supports this hypothesis: ANOVA tests confirm a significant effect for those who received help ($F=26.9$ and $\alpha=3.314e-07$ -- an increase of two times more shared onions per soup delivered) compared to those who did not receive help ($F=0.6$ and $\alpha=0.44$). These results show that our study design captured altruistic behavior differences (Cohen $d = 0.28$ when comparing Rounds 1 and 3).

While the remainder of this article relies on the fact that the two cultural groups in our dataset demonstrated different altruistic behaviors in Round 1, we note that we also found  two other characteristics influencing behavior that may be worth investigating in future studies: gender and political leaning (see Supplementary Material).

\subsection*{Inverse Reinforcement Learning of Culturally-Attuned Human Values}
\begin{figure*}[t]
    \centering
    \begin{subfigure}[b]{0.475\textwidth}
        \centering
        \includegraphics[width=\textwidth]{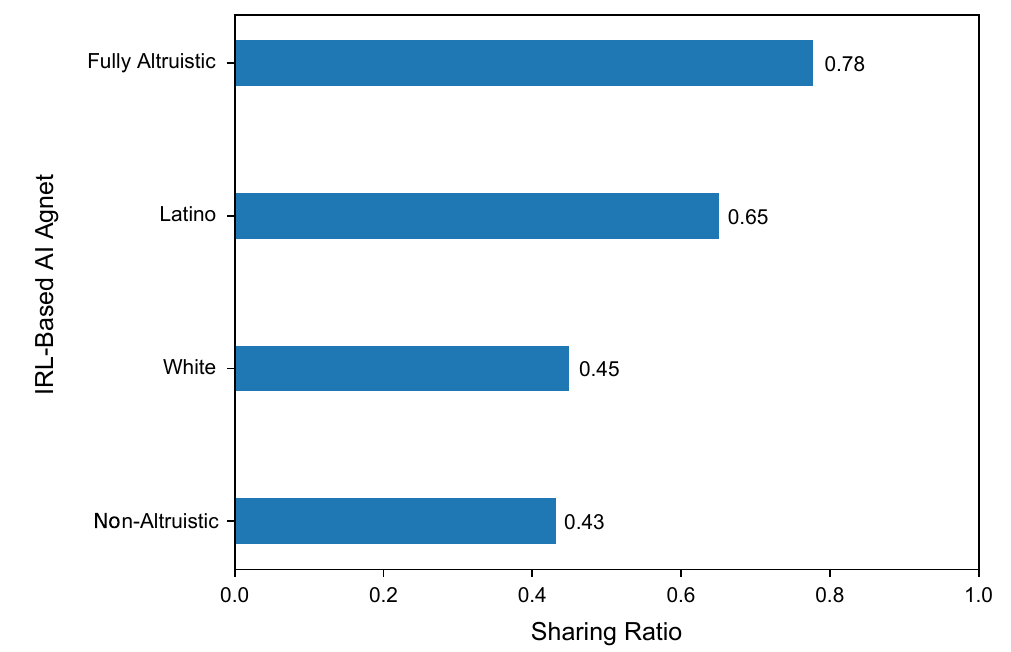}
        \caption[]%
        {}
        \label{fig:results:groups}
    \end{subfigure}
    \hfill
    \begin{subfigure}[b]{0.475\textwidth}  
        \centering 
        \includegraphics[width=\textwidth]{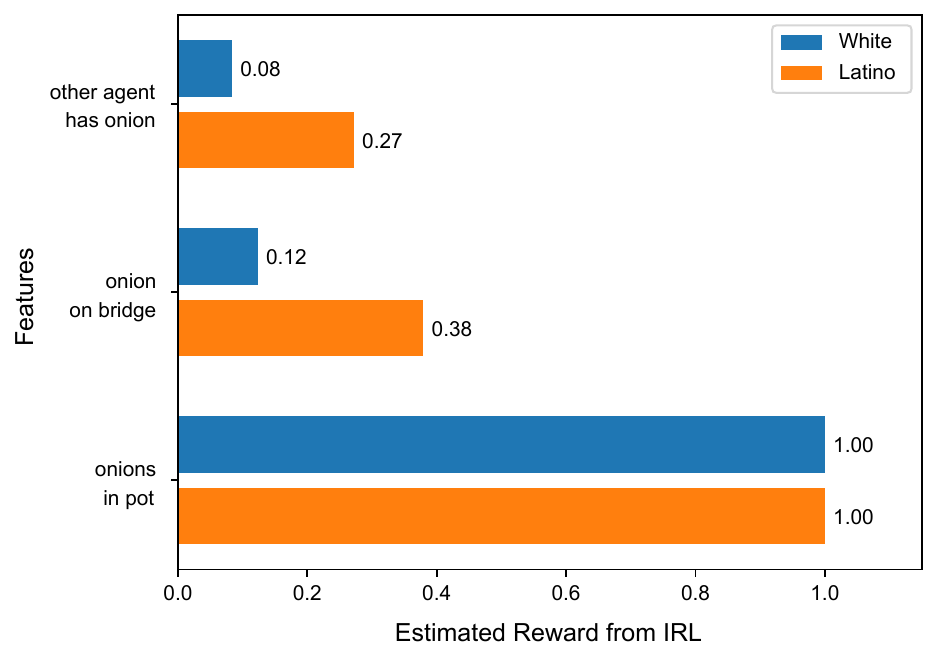}
        \caption[]%
        {}    
        \label{fig:results:features}
    \end{subfigure}
 %   \vskip\baselineskip
    \caption[]{{\bf Inverse Reinforcement Learning (IRL) based Analysis of Altruistic Behaviors in the Online Experiment.} {\bf (a)} The bars represent the Sharing Ratios computed from the behaviors (trajectories in the game) for the four different IRL-based AI agents on the Y axis (see text for details). {\bf (b)} The scaled reward values assigned to three different features extracted from the game, showing how the selected features contribute to the final reward value estimated for White versus Latino participants.}
    %\label{fig:results}
\end{figure*}

We next tested our hypothesis that AI agents could learn culture-specific values and norms by observing and learning from the actions of humans from particular cultural groups. Specifically, we used inverse reinforcement learning (IRL) to recover the underlying reward functions for participants who identified themselves as Latino or White participants. We then tested whether such an IRL-based AI can (1) generate behavior aligned with the norms of the cultural group it was trained on, and (2) also rely on the same learned reward function to generalize and make decisions when confronted with novel altruistic decision making scenarios (here, previously unseen variations of the game).

We quantified the altruistic tendency of IRL-based AI agents trained on a particular group's data using the \textit{Sharing Ratio (SR)} measure, calculated as the {\em average reward of the sharing trajectory ($ART_S$)} divided by the {\em average reward of the cooking trajectory ($ART_C$)} where the average reward of a trajectory (ART) is defined as:

\begin{equation}
\label{eq:average_reward_trajectory}
ART(\tau,rf) = [\sum^{n_\tau}_{t=1}{norm(rf(t))}]/{n_\tau}
\end{equation}
where $\tau$ is a trajectory (for cooking or sharing behavior) with $n_\tau$ time steps, $norm$ is the min-max normalization applied per trajectory step, and $rf$ is the learned reward function for each of the four computed IRL models.  

Defined as above, the SR for a given IRL model indicates how much more rewarding a group considers sharing an onion compared to using it for cooking their own soups: the higher the sharing ratio, the more altruistic the AI agent behaves.
Figure~\ref{fig:results:groups} compares the Sharing Ratio (SR) for the two IRL-based AI agents trained on the Latino and White American participant groups, with two other agents trained on datasets for the two extreme behaviors, namely, a fully altruistic agent who shares every onion, and a non-altruistic agent who does not share onions. As expected, the fully altruistic agent and the non-altruistic agent had the highest and lowest sharing ratios, respectively.
The sharing ratio for agents trained on Latino participants was closer to the altruistic agent's sharing ratio and higher than the ratio for White participants, which in turn was higher than the ratio for the non-altruistic agent  (Figure~\ref{fig:results:groups}). These results indicate that the learned reward functions for the Latino and White American groups are consistent with the human behavioral tendencies reported in the previous section. They additionally help to quantify the extent to which the two groups of participants exhibit altruistic behavior in the game. 

\begin{figure}
    \centering
    \subfloat[original layout]{\includegraphics[width=0.35\textwidth]{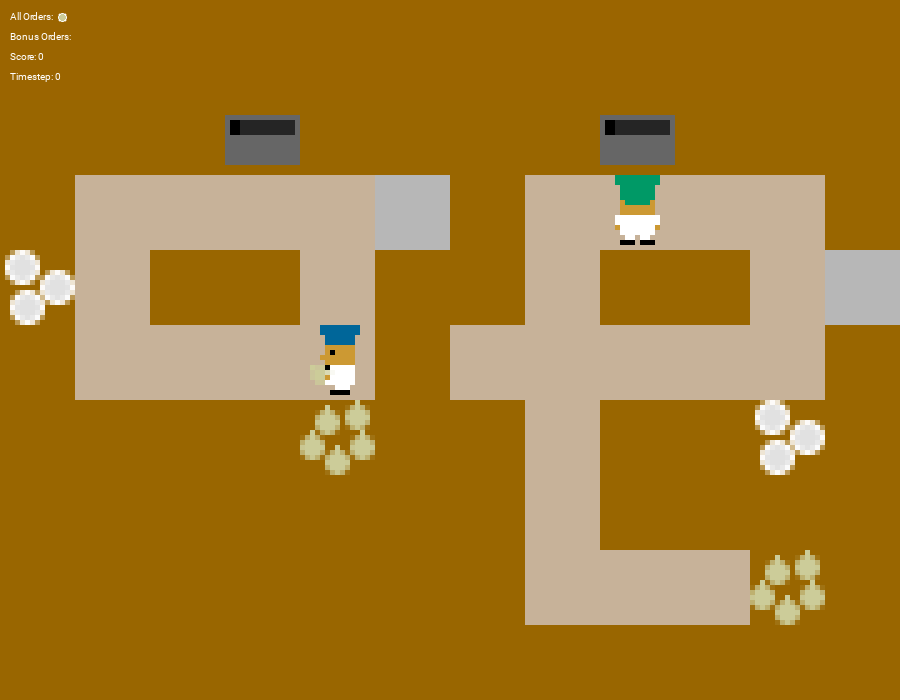}}

    \medskip % Adjust the vertical space between rows
    
    \subfloat[modified layout 1]{\includegraphics[width=0.235\textwidth]{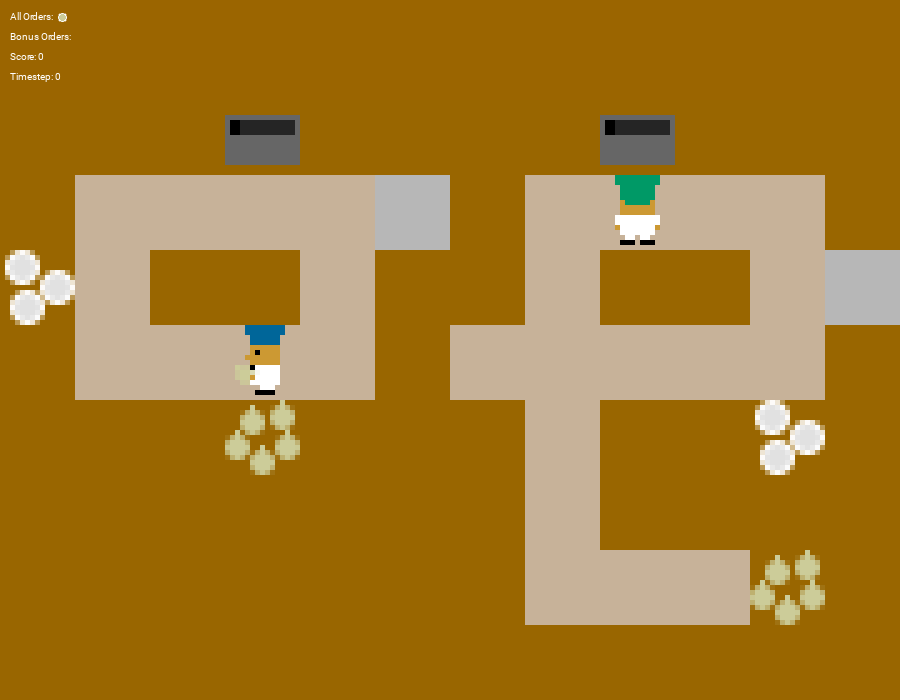}}\hfill
    \subfloat[modified layout 2]{\includegraphics[width=0.235\textwidth]{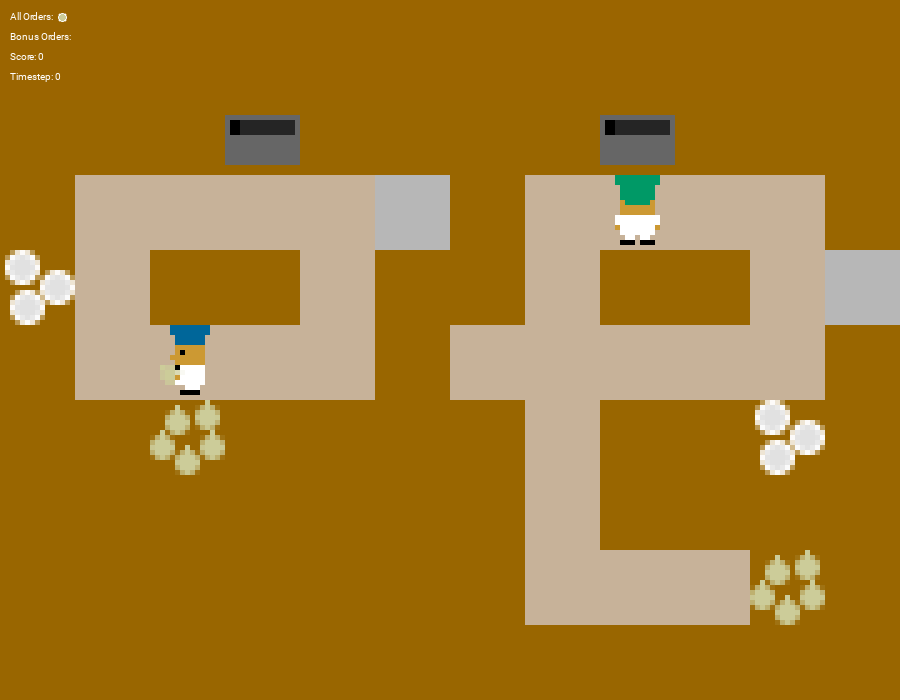}}\hfill

        \medskip % Adjust the vertical space between rows
    \subfloat[modified layout 3]{\includegraphics[width=0.235\textwidth]{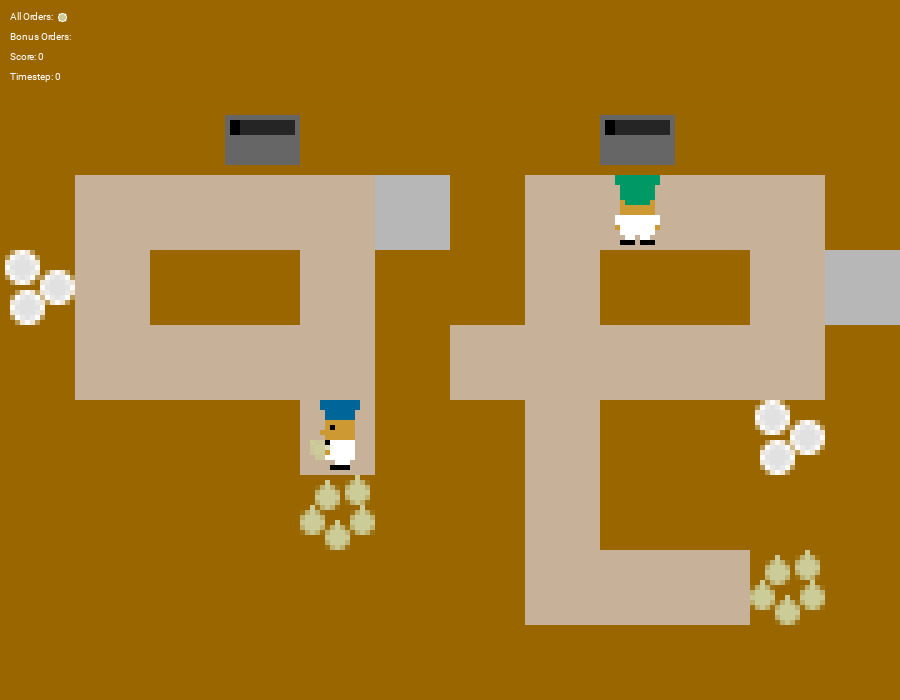}}
    \hfill 
    \subfloat[modified layout 4]{\includegraphics[width=0.235\textwidth]{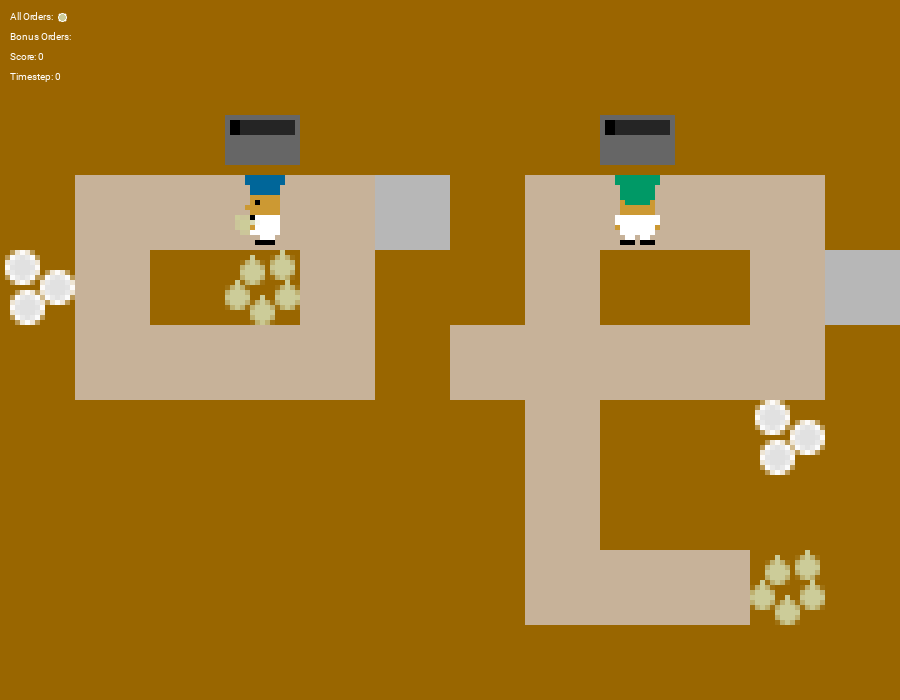}}\hfill
            \medskip % Adjust the vertical space between rows
    \subfloat[modified layout 5]{\includegraphics[width=0.235\textwidth]{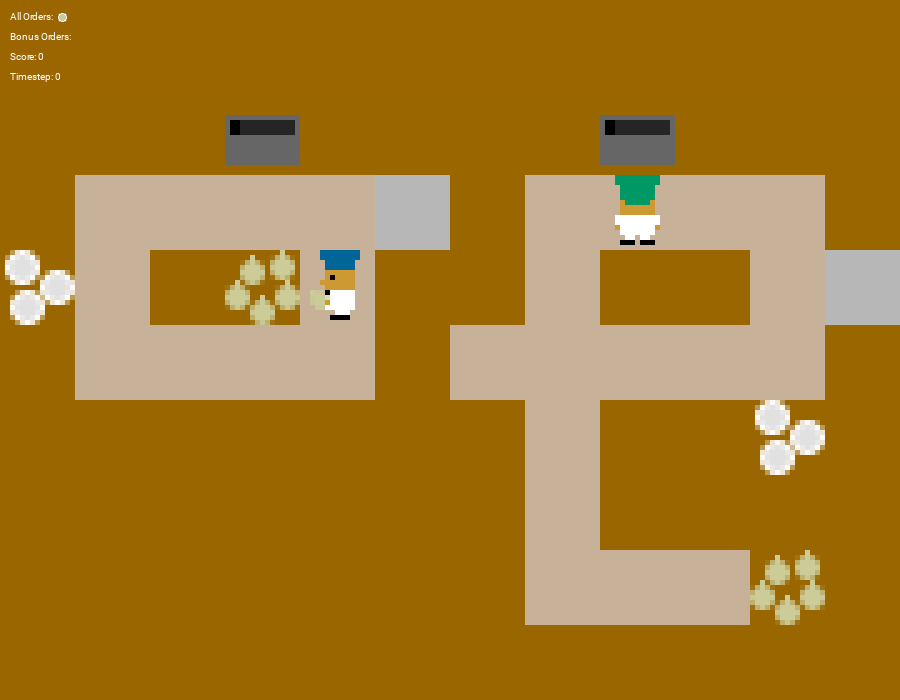}}\hfill
    \subfloat[modified layout 6]{\includegraphics[width=0.235\textwidth]{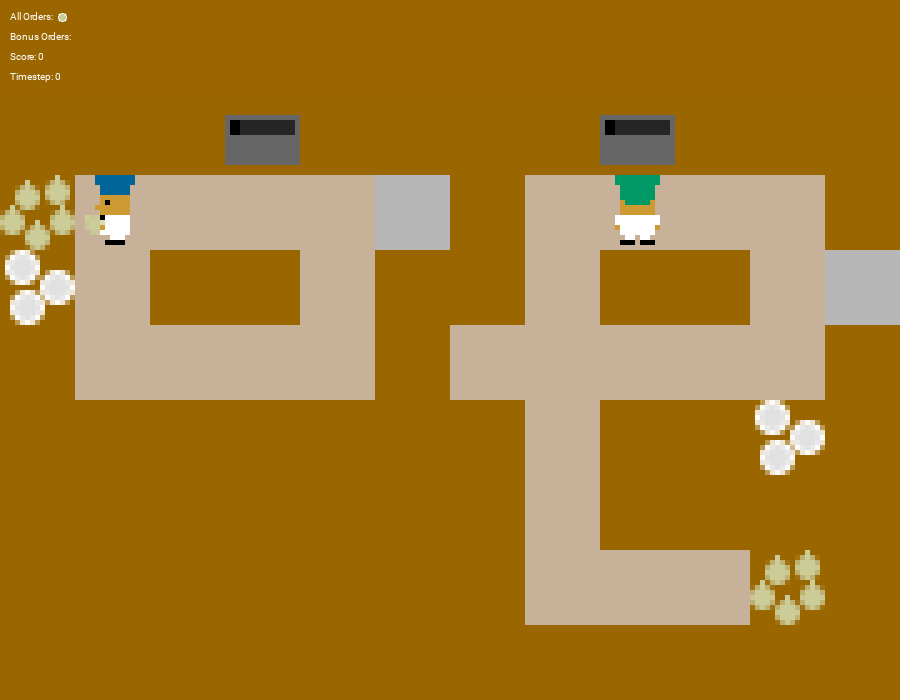}}

    \caption{{\bf Testing the Generalizability of the IRL-Based Reward Function}. By modifying the original layout shown in {\bf (a)}, we created six new IRL environments by changing the player position, onion store position, and counter layout. The idea behind the changes was to alter the level of effort needed to share an onion. In the new layouts {\bf (b)} through {\bf (d)}, sharing an onion requires less steps than cooking. Conversely, modified layouts {\bf (e)} and {\bf (g)} require more steps to render help than to cook, while modified layout {\bf (f)} requires approximately the same number of steps for both helping and cooking.}
    \label{fig:layouts}
\end{figure}

\subsubsection*{Explainable AI}
Besides estimating an intrinsic reward function for each cultural group in the game, the IRL models offer a unique opportunity to analyze how this knowledge is expressed in terms of different features of the task, in this case, features of the Overcooked game. This aspect of our approach to learning value systems for AI agents fits well with the growing realization that deployed AI systems need to be explainable \cite{Muller-explainable-AI-2019,Linardatos-XAI-2021} rather than black-box models. 

Figure~\ref{fig:results:features} shows the valuation of three of the most relevant features when comparing the Latino versus White cultural groups in our dataset. The fact that both groups highly valued having onions in the pot captures the explicit goal of the game (bottom two bars in the plot). Besides the common goal of cooking more soups, the reward function also captures the differences between the two cultural groups. The observation that the Latino participants provided more help to the other player is captured by another feature in  Figure~\ref{fig:results:features}: whether there is an onion on the cooperation bridge, which has a higher value for Latino participants than White participants. The third feature in  Figure~\ref{fig:results:features} -- whether the other agent is holding an onion -- is also directly related to a participant's sense of fairness and altruistic behavior. The learned reward value for this feature is higher for Latino participants than White participants, consistent with the overall differences in altruistic behavior for these two cultural groups seen in our data.

\subsubsection*{Generalizability of the Learned Reward Function}
\begin{figure}
    \centering
    \includegraphics[width=0.95\linewidth]{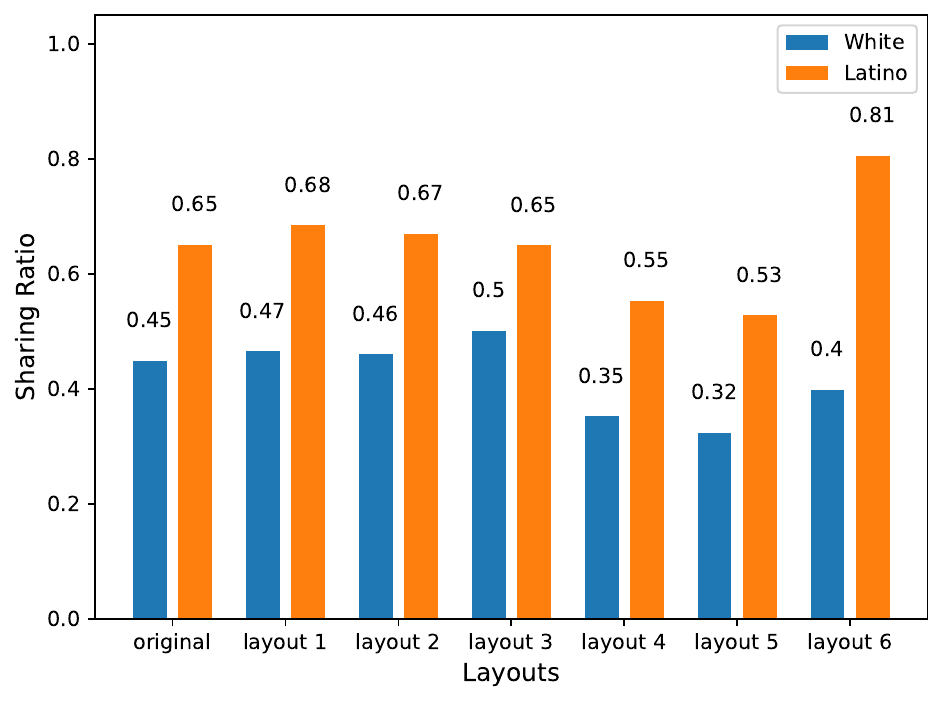}
    \caption{{\bf Comparison of Altruistic Behavior of Culturally-Attuned AI Agents in the Original and Six Novel Environments}. Each bar represents the Sharing Ratio (SR) for an AI agent trained on the behavior of White or Latino participants. A higher ratio indicates a preference for helping the other player than cooking one's own soup, based on the reward function learned from humans playing in the original game layout. The sharing ratio of the IRL agent trained on data from Latino participants was higher than the IRL agent trained on White participants in all layouts. In both the original layout and layouts 1 through 3, where it requires fewer steps to help than to cook, both IRL agents exhibit relatively high sharing ratios.}
    \label{fig:layout-results}
\end{figure}

To demonstrate how our proposed IRL-based approach can learn a generalizable culturally-learned behavior, we examined how the reward function learned in the original game layout would perform in six other novel layouts with no further training (see Figure~\ref{fig:layouts}). We summarize our findings in Figure~\ref{fig:layout-results}, which fits well with our expectation that an agent using the reward function learned from a certain cultural group will exhibit a behavior consistent with the expected altruistic values of that group.

\section*{Discussion}
Given the rapid strides being made in AI research today, a question of utmost importance is how AI systems can be imbued with human values and morals. Rather than attempting to hardwire a universal moral code or learn a moral code from large datasets of precompiled human judgments, we propose that AI agents should be designed to learn culturally-attuned moral codes and do so by emulating how humans learn such codes: from observation and interactions with other humans within the community or culture the AI has been deployed in. To illustrate this approach, we presented the first proof-of-concept results from an online experiment involving 300 subjects. Our results demonstrate how inverse reinforcement learning (IRL) can be used by AI agents to learn culture-specific reward functions from the behavior of humans from a particular cultural background (here, Latino or White ethnic backgrounds). Specifically, an AI agent that was trained on data from Latino American  participants replicated the high level of altruism it had learned from this specific human cultural group, even in a new context.

We also explored how the demonstration of altruistic behavior by another player in one round of interaction can trigger altruistic behavior in the next round, akin to behavioral accommodation in which people adjust their behavior to that of others. Our results  show that the degree of such behavioral accommodation may vary according to cultural background~\cite{bi_liked_2014}. 

In contrast to other approaches based on deep learning and large language models, the reward function learned by the AI agent from human behavior offers interpretability through  analysis of an IRL reward function's features. Furthermore, the features-based reward function learned in one task can generalize to other tasks, illustrating the utility of an IRL-based approach to learning culturally-attuned moral codes.

Our results leave open several important directions that are worthy of further investigation. First, our proof-of-concept demonstration only involved two cultural groups. The AI agents were trained separately for the two groups, learning from the average behavior of each group. It would be interesting to deploy AI agents {\em de novo} in different communities, nations, and other cultural groups and analyze the resulting reward functions to understand what type of moral code or norm is implicitly recovered. Second, the features used in our IRL implementation were engineered for the Overcooked environment; we hope to endow future IRL-based AI agents with feature-learning abilities to customize features to each environment while retaining interpretability. Finally, our online experiment involved a very limited type of altruistic behavior in an online game setting; the IRL approach is much more general and could, in principle, learn complex multi-dimensional reward functions quantifying moral and ethical codes governing human behavior within a culture. Exploring how such reward functions can be learned from large-scale complex human interactions is an important direction of future research aimed at understanding how—like a child—an AI can learn and enact the cultural patterns and value systems to which it is exposed during its development. 

\section*{Methods}

\subsection*{Dataset}
Our sample of N=300 participants was estimated based on a power analysis ($\alpha=0.05$ and $\beta=.10$) considering the altruistic effect across participants' self-declared ethnic background of $f=0.18$ (Cohen $f$). This effect was calculated based on the initial sample of 100 participants. Two-thirds of our participants (N=203) were recruited through the Prolific platform. Due to institutional restrictions, we recruited the second third of our dataset using the Positly (N=59) and the MTurk (N=38) platforms. Despite this difference, the participants' experience with the study was exactly the same on all platforms, as they were redirected to our website when they accepted the offer. We prioritized Prolific because of its advanced population qualifiers, which allowed for a more precise pre-screening of participants who were paid US\$ 2 for an average of 10 minutes of their time. Table~\ref{tab:demographics} presents the demographic characteristics of the sample data we collected as the first step of our study.

\begin{table}[h]
    \centering
    \begin{tabular}{|c|c|c|c|}
    \hline
        \textbf{Gender} & \textbf{Political} & \textbf{Race} & \textbf{Age} \\
    \hline
        Female: 205 & Center: 101 & Latino: 110 & mean: 30.6 \\
        Male: 95 & Left: 141 & White: 190 & sd: 10.4 \\
                & Right: 58 &           & median: 28 \\
    \hline
    \end{tabular}
    \vspace{0.1in}
    \caption{Summary of the demographic characteristics of the 300 participants in our dataset.}
    \label{tab:demographics}
\end{table}

After responding to the demographics questionnaire, participants engaged in three training steps on the game mechanics. Participants could progress only if they demonstrated learning the basic actions of the game: move around the kitchen, move ingredients to the stove location, and deliver a soup. In fact, all participants were able to either deliver a soup or provide help at least once through the three rounds of the game, with 90\% of them delivering soup or providing help at least four times and half of them doing so more than ten times.

\subsection*{Model Training}

We used data from the first round of the game as the training dataset for IRL. Assuming that the participant's behavior after picking up an onion (either to give the onion to the other player or to use it to cook one’s own soup) best represents the difference between the two cultural groups, we extracted these ``onion-delivering'' traces from the 300 game trajectories. A trace starts with the participant picking up the onion and ends with dropping off the onion at the `cooperation bridge' or the player's own pot. Each trace is labeled as `altruistic' or `non-altruistic' depending on the player's placement of the onion. Traces were also labeled according to whether the trace is from a player with a Latino or White American background. We finally compacted the traces by removing states where the participant-controlled agent did not move. This resulted in a total of 2958 traces, out of which 476 were labeled as ``altruistic.'' The traces were then grouped into four different training datasets based on their labels: an altruistic dataset, a non-altruistic dataset, a Latino American dataset, and a White American dataset.

\subsection*{IRL}
We used the Maximum Entropy Deep Inverse Reinforcement Learning algorithm~\cite{wulfmier_maxentirl_2015} paired with the Population Based Training algorithm (PBT)~\cite{Jaderberg-PBT-2017} and Proximal Policy Optimization (PPO)~\cite{Schulman-PPO-2017} for the reinforcement learning algorithm. The reward function was implemented using a neural network architecture comprising two layers, a linear input layer to transform an input vector into a hidden vector of size 200. The Exponential Linear Unit (ELU) activation function was then applied to this hidden vector. A linear output layer then converted the hidden layer activity into a scalar reward value.  

A set of features, chosen based on heuristics (see Table~\ref{tab:feature_vector}), was extracted from each game state as the input to the reward function. This feature vector has a size of 18. The training objective was to learn a reward function such that the RL policy maximizing expected reward according to the learned reward function imitates the human behavior in the data used for training. Thus, we evaluated the reward function by comparing the similarity between the learned policy and human behavior (using mean squared error between trajectories generated by the policy and humans). 

\begin{table*}
    \centering
    \begin{tabular}{|c|c|c|c|}
        \hline
        \textbf{Features} & \textbf{Vector Size} & \textbf{Features} & \textbf{Vector Size} \\
        \hline
        agent's position relative to onion store & 2 & onion on bridge & 1 \\
        \hline
        agent's position relative to bridge & 2 & onions in pot & 1 \\
        \hline
        agent's position relative to stove & 2 & agent has onion & 1 \\
        \hline
        agent's orientation & 4 & other agent has onion & 1 \\
        \hline
        agent on shortest path from starting  &  &  &  \\
        position to stove with onion in hand& 4  &  &  \\
        \hline
    \end{tabular}
     \vspace{0.1in}
    \caption{Features used for IRL with their vector sizes. Here, ``agent'' refers to the AI-controlled agent operating on the left side in  the original game layout (Figure~\ref{fig:game}) and ``other agent'' refers to the agent on the right side. Features were chosen and fine-tuned based on heuristics and training results.}
    \label{tab:feature_vector}
\end{table*}

To train the reward function, we used an SGD optimizer with a learning rate of 0.001 and weight decay of 0.9. We used an exponential LR scheduler with gamma = 0.999. The start state of the training is shown in Figure~\ref{fig:layouts} for the different layouts. The blue agent (on the left side) starts at the position shown and needs to decide whether to take actions to share the onion by placing the onion on the bridge or to place the onion in the pot. We trained four reward functions using the four types of behaviors generated by the altruistic bot, non-altruistic bot, Latino participants, and White participants (see main text for details).

\acknow{This material is based upon work supported by the Templeton World Charity Foundation, the National Science Foundation (NSF) EFRI Grant no.\ 2223495, a UW + Amazon Science Hub grant, and a Cherng Jia \& Elizabeth Yun Hwang Professorship to RPNR. The opinions expressed in this publication are those of the authors and do not necessarily reflect the views of the funders.
}

\showacknow{} % Display the acknowledgments section

\bibliography{moralai.bib}

\newpage
\makeatletter
\renewcommand \thesection{S\@arabic\c@section}
\renewcommand\thetable{S\@arabic\c@table}
\renewcommand \thealgorithm{S\@arabic\c@algorithm}
\renewcommand \thefigure{S\@arabic\c@figure}
\setcounter{figure}{0}
\makeatother
\section*{Supplementary Material}
\subsection*{Influence of Other Demographic Characteristics}
\begin{figure*}[h]
    \centering
    \begin{subfigure}[b]{0.48\linewidth}
        \includegraphics[width=\linewidth]{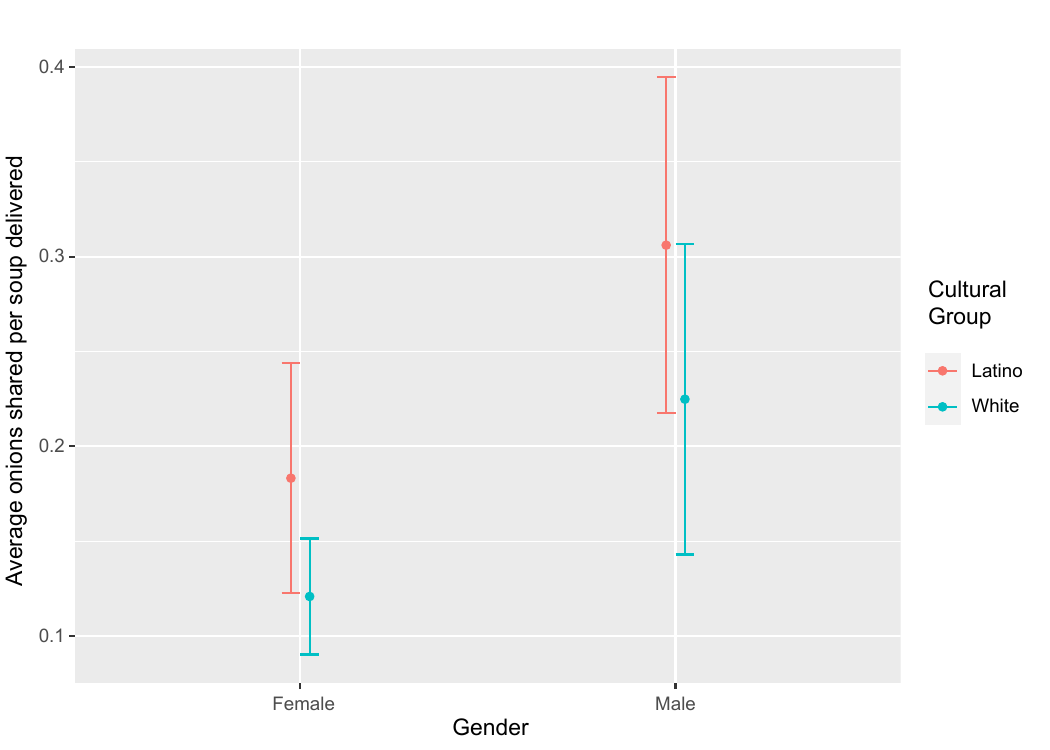}
        \caption[]%
        {}
        \label{fig:gender:race}
    \end{subfigure}
    \begin{subfigure}[b]{0.48\linewidth}
        \includegraphics[width=\linewidth]{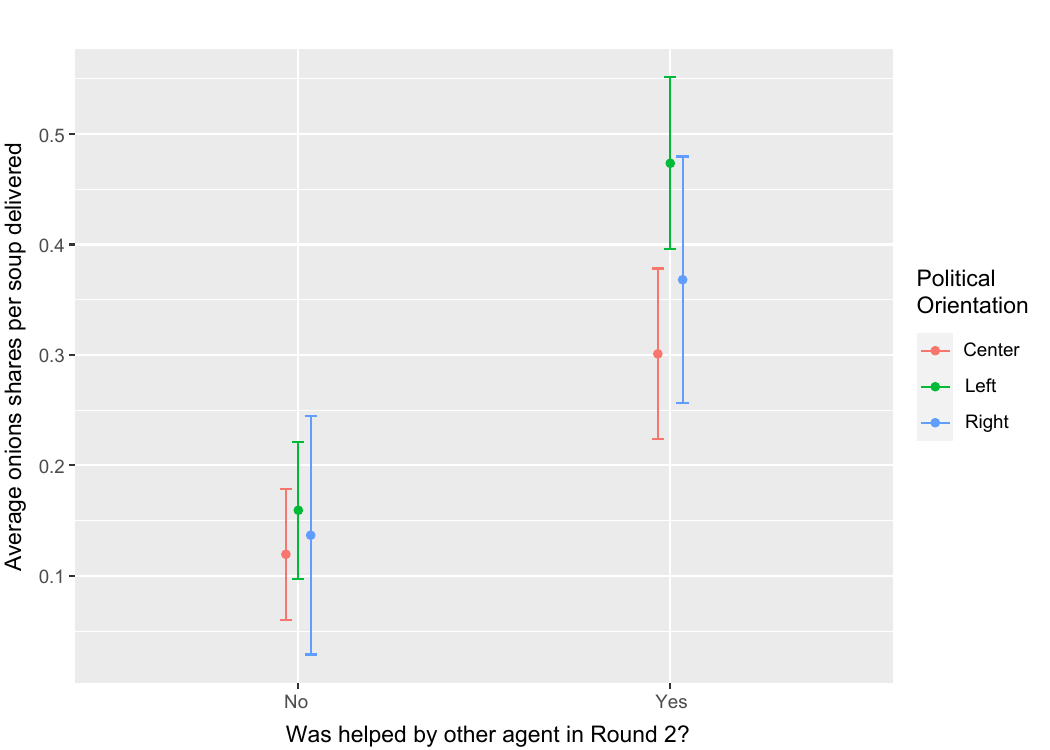}
        \caption[]%
        {}
        \label{fig:political:coop}
    \end{subfigure}

    %\vskip\baselineskip
    \caption[]{{\bf Influence of Other Demographic Characteristics on Altruistic Behavior in the Online Experiment}. {\bf (a)} Influence of gender. {\bf (b)} Influence of political orientation. } 
    \label{fig:results2}
\end{figure*}

During our analysis of altruistic behaviors in our data, in addition to the influence of cultural background (Latino versus White; see main text), we found two other demographic  characteristics influencing behavior that may be worthy of investigation in future studies: 
\begin{itemize}
\item \textit{Gender:} While controlling for age, gender, and political orientation in the data from our online experiment, we found that male participants tended to be more atruistic (about 10\% more) than female players in Round 1 (Figure~\ref{fig:gender:race}). Although this effect was double the size of the one identified overall when comparing the two cultural groups, we confirmed that the Latino participants as a group were significantly more collaborative than the White participants. 
\item \textit{Political Orientation:} 
We also found a relation between experimental treatment in Round 2 and political orientation. As shown in Figure~\ref{fig:political:coop}, those who received help in Round 2, and who held left-leaning political views, showed more altruism in Round 3 compared to those with centrist political views. This relation was not evidenced among those who did not receive help, i.e., their altruism in Round 3 was untethered from their political ideology.\footnote{A power analysis of the effect of helping behavior across the three political-leaning groups (Cohen $f=0.14$) shows a need for larger sample size ($n=205$ per group) to rule out any false negative errors.}
\end{itemize}

\subsection*{Example Trajectories of IRL-based AI Agents}
\begin{figure*}
    \centering
    \begin{subfigure}[b]{0.3\linewidth}
        \includegraphics[width=\linewidth]{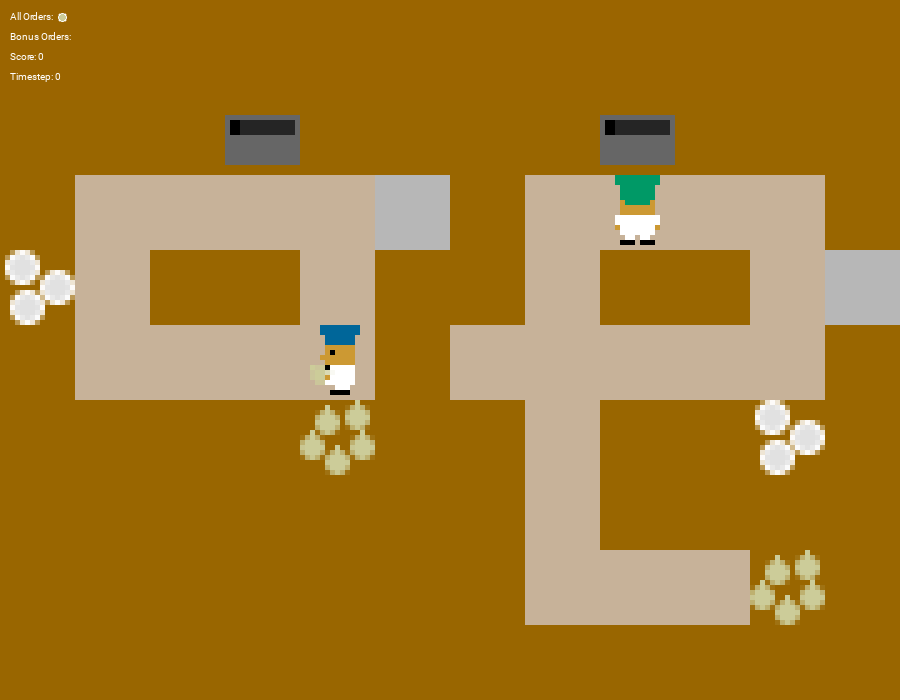}
    \end{subfigure}
    \begin{subfigure}[b]{0.3\linewidth}
        \includegraphics[width=\linewidth]{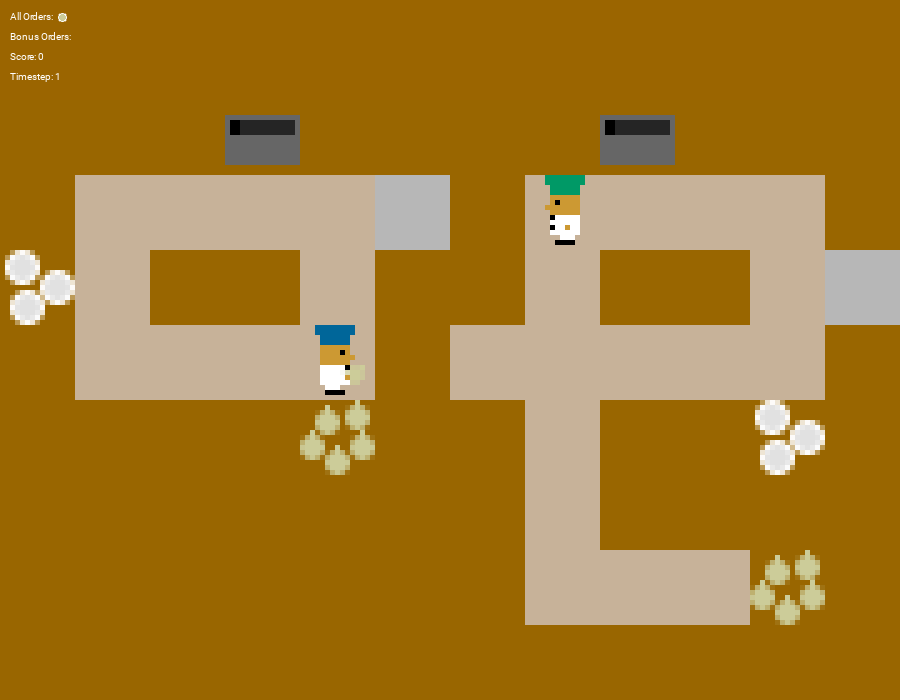}
    \end{subfigure}
    \begin{subfigure}[b]{0.3\linewidth}
        \includegraphics[width=\linewidth]{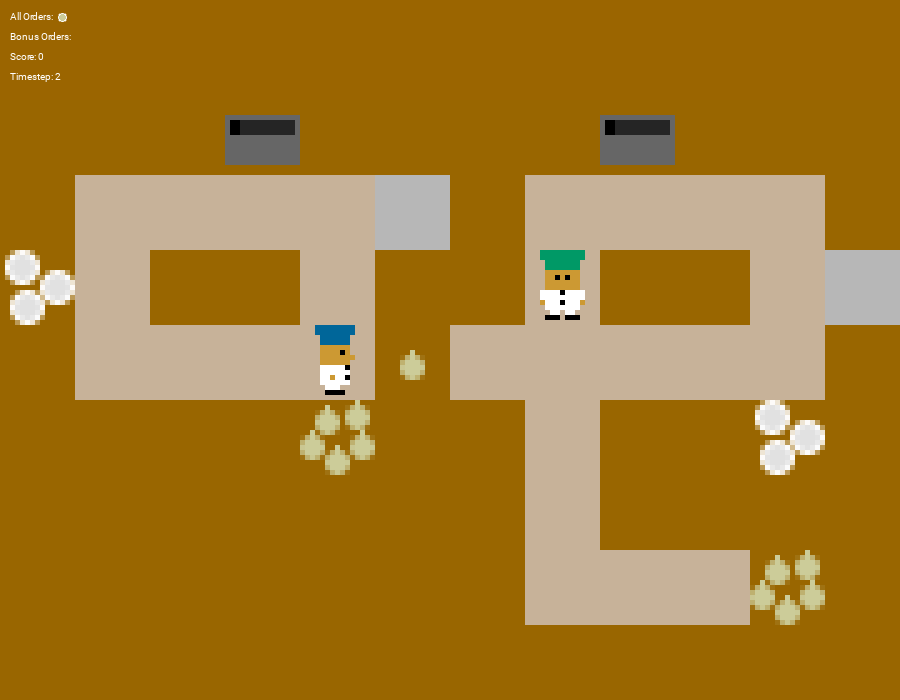}
    \end{subfigure}
    
    \caption{{\bf Example of altruistic behavior by an IRL-based AI agent trained using data from Latino participants in the online experiment}. This trajectory (in the original game layout and read from left to right) illustrates how the trained agent (with blue hat) has learned, as part of its policy, to turn towards the cooperation bridge and provide assistance to the other agent by placing an onion on the bridge.}
    \label{fig:suppl-1}
\end{figure*}

\begin{figure*}
    \centering
    \begin{subfigure}[b]{0.3\linewidth}
        \includegraphics[width=\linewidth]{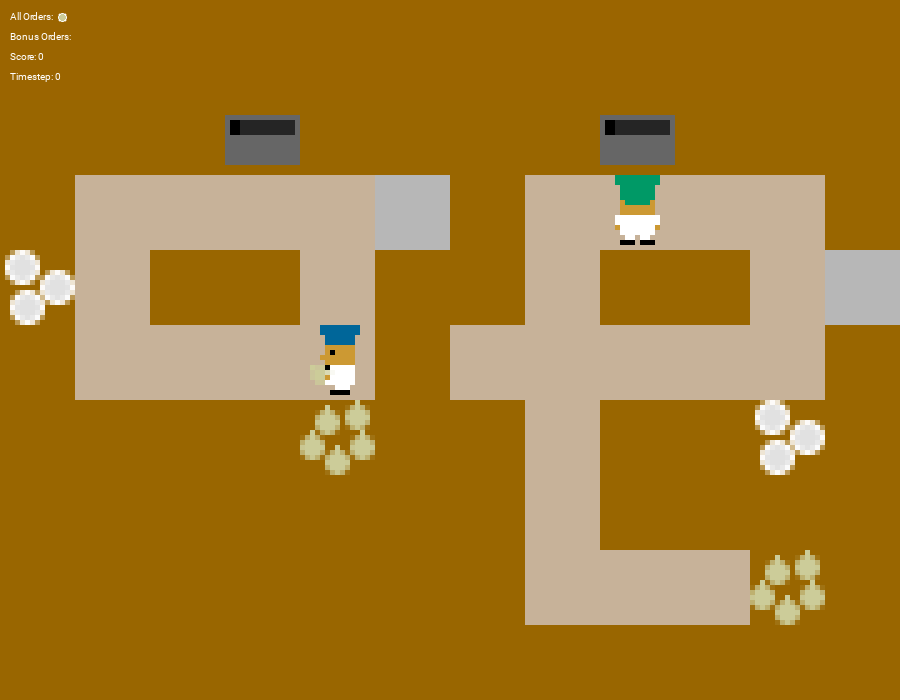}
    \end{subfigure}
    \begin{subfigure}[b]{0.3\linewidth}
        \includegraphics[width=\linewidth]{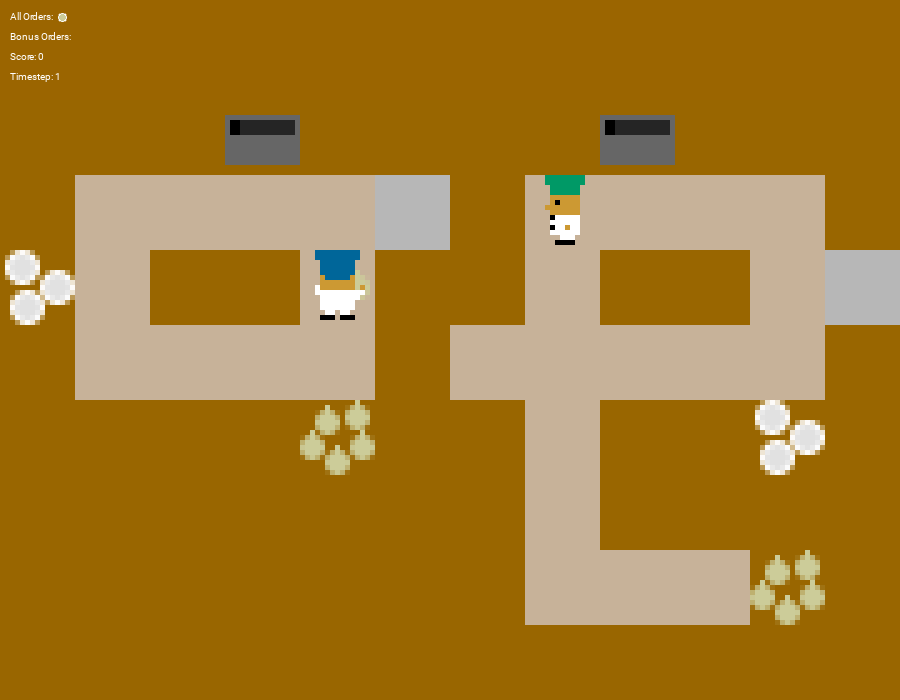}
    \end{subfigure}
    \begin{subfigure}[b]{0.3\linewidth}
        \includegraphics[width=\linewidth]{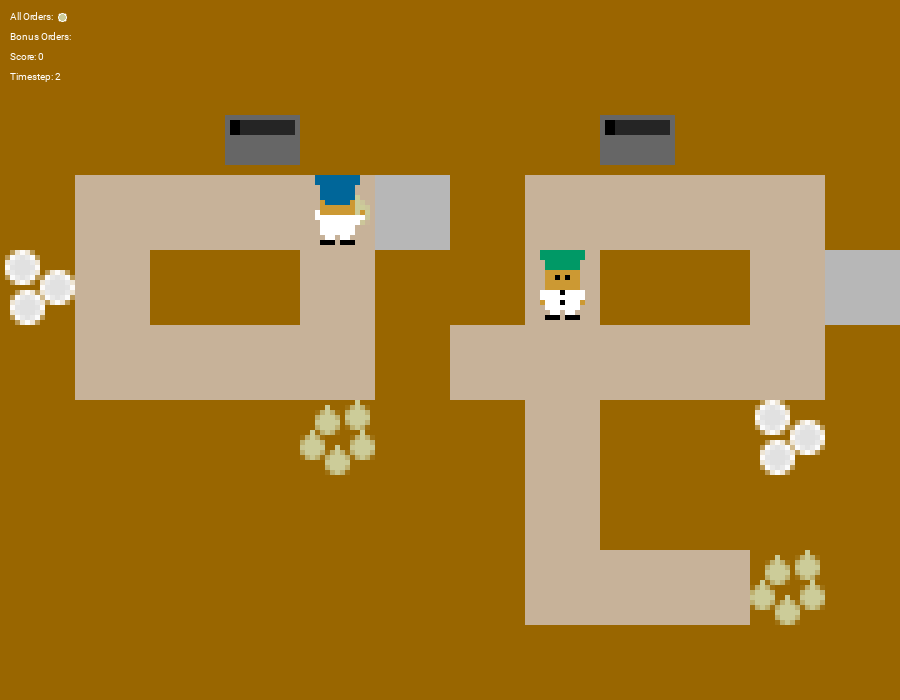}
    \end{subfigure}
    
    \begin{subfigure}[b]{0.3\linewidth}
        \includegraphics[width=\linewidth]{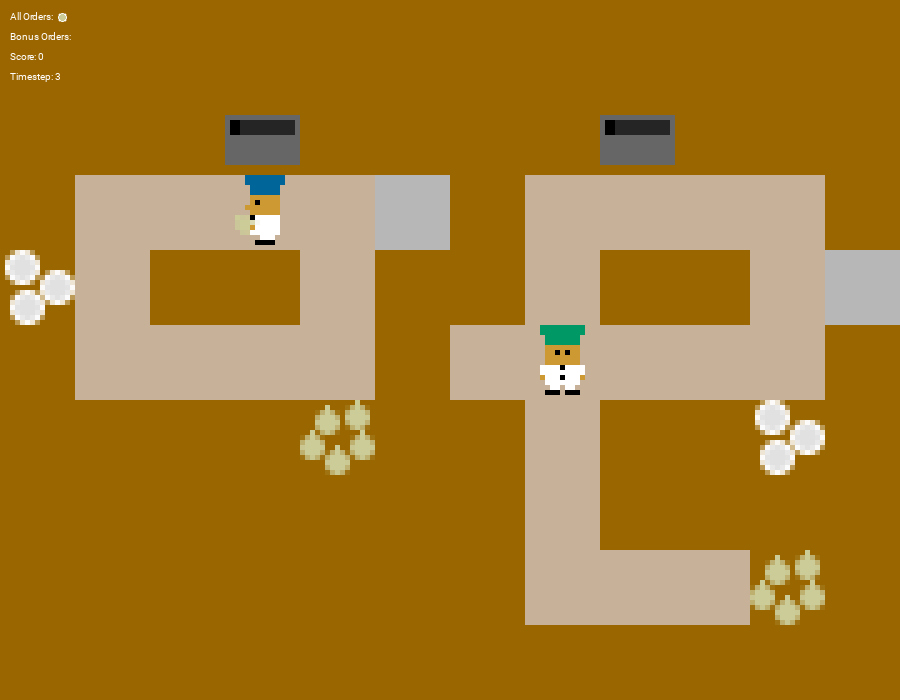}
    \end{subfigure}
    \begin{subfigure}[b]{0.3\linewidth}
        \includegraphics[width=\linewidth]{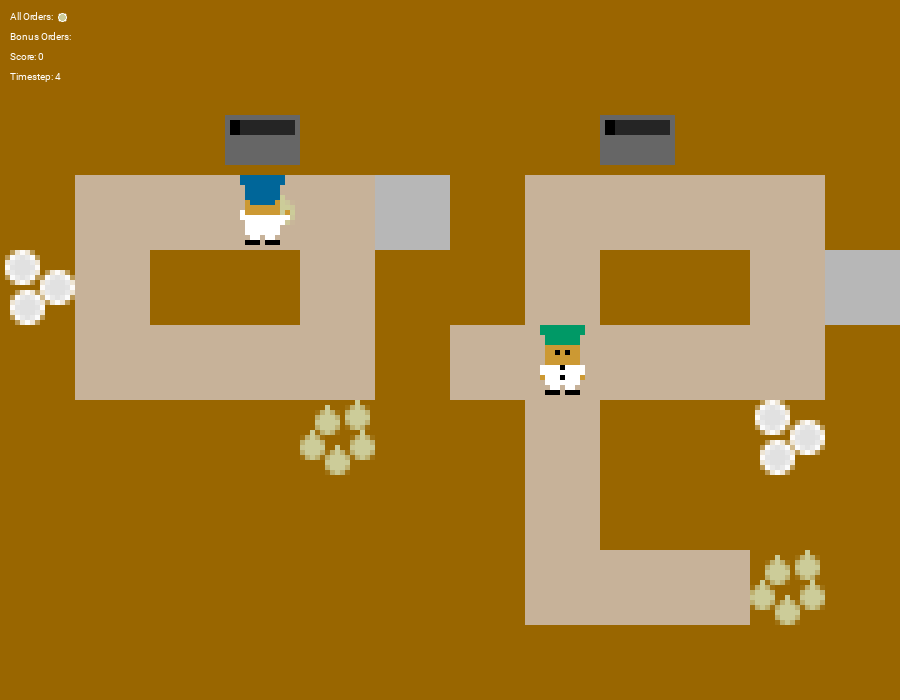}
    \end{subfigure}
    \begin{subfigure}[b]{0.3\linewidth}
        \includegraphics[width=\linewidth]{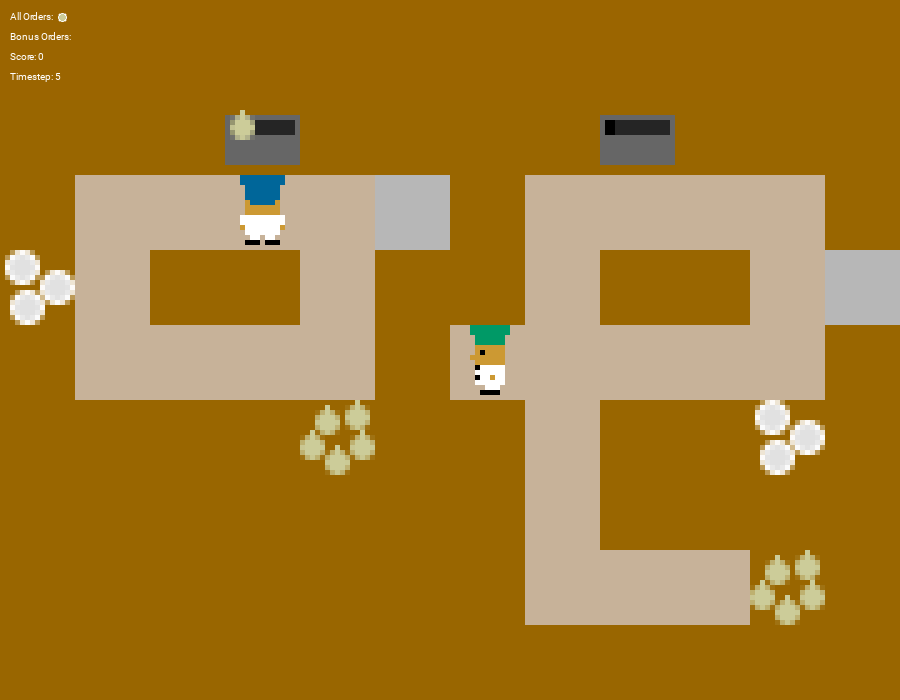}
    \end{subfigure}
    
    \caption{{\bf Example of non-altruistic behavior by an IRL-based AI agent trained using data from White participants in the online experiment}. This trajectory (in the original game layout) illustrates how the trained agent (with blue hat) has learned, as part of its policy, to navigate to the stove (rather than the cooperation bridge) to place the onion in the pot for cooking its own soup. The trajectory is read from left to right, top to bottom.}
    \label{fig:suppl-2}
\end{figure*}

Figures~\ref{fig:suppl-1} and \ref{fig:suppl-2} show snapshots from example trajectories of two AI agents trained using IRL based on data from Latino and White participants, respectively. The first trajectory illustrates altruistic behavior (sharing an onion) while the second illustrates non-altruistic behavior (using the onion to cook its own soup).) 

\end{document}